\documentclass{article}

\usepackage[final]{neurips_2019}
\usepackage[utf8]{inputenc} % allow utf-8 input
\usepackage[T1]{fontenc}    % use 8-bit T1 fonts
\usepackage{hyperref}       % hyperlinks
\usepackage{url}            % simple URL type setting
\usepackage{booktabs}       % professional-quality tables
\usepackage{amsfonts}       % blackboard math symbols
\usepackage{nicefrac}       % compact symbols for 1/2, etc.
\usepackage{microtype}      % microtypography
\usepackage[most]{tcolorbox}
\usepackage{subcaption}
\usepackage{sidecap}
\usepackage{wrapfig}
\usepackage{lipsum}

\usepackage{graphicx}

\title{A Weak Supervision Approach to Detecting Visual Anomalies for Automated Testing of Graphics Units}

\author{%
  Adi Szeskin \\
  Intel, IT Advanced Analytics\\
  \texttt{adi.szeskin@intel.com} \\
  \And
   Lev Faivishevsky \\
  Intel, IT Advanced Analytics\\
  \texttt{lev.faivishevsky@intel.com} \\
  \And
   Ashwin K Muppalla \\
  Intel, Visual Processing Group\\
  \texttt{ashwin.muppalla@intel.com} \\
  \And
   Amitai Armon \\
  Intel, IT Advanced Analytics\\
  \texttt{amitai.armon@intel.com} \\
  \And
   Tom Hope \\
  Intel, IT Advanced Analytics\\
  \texttt{tom.hope@intel.com} \\
}

\begin{document}

\maketitle

\begin{abstract}
We present a deep learning system for testing graphics units by detecting novel visual corruptions in videos. Unlike previous work in which manual tagging was required to collect labeled training data, our weak supervision method is fully automatic and needs no human labelling. This is achieved by reproducing driver bugs that increase the \textit{probability} of generating corruptions, and by making use of ideas and methods from the Multiple Instance Learning (MIL) setting.
In our experiments, we significantly outperform unsupervised methods such as GAN-based models and discover novel corruptions undetected by baselines, while adhering to strict requirements on accuracy and efficiency of our real-time system. 

\end{abstract}

\section{Introduction}
\label{intro-sec}

Graphics processing units (GPUs), complex pieces of hardware responsible for rendering images for display, are essential components in personal computers, mobile phones, embedded systems and gaming consoles. The complexity of modern GPUs mandates validation and testing for defects, checking both hardware and software (e.g., drivers) with diverse systems (e.g., screens) and content.

%These visual anomalies typically map to known common types.

\begin{figure}[]
\centering
    %\centerline{\includegraphics[height = 8cm]{figures/System4}}
    \includegraphics[height=0.2\textwidth, width=0.3\textwidth]{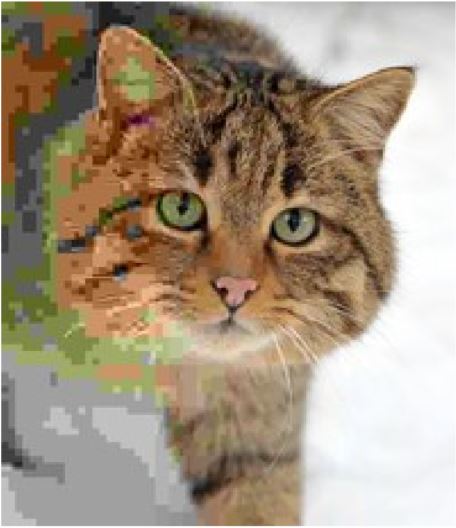}
    \includegraphics[height=0.2\textwidth, width=0.3\textwidth]{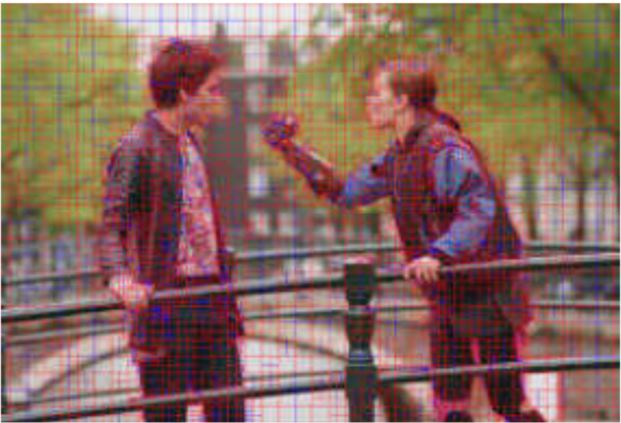}
    \includegraphics[height=0.2\textwidth, width=0.3\textwidth]{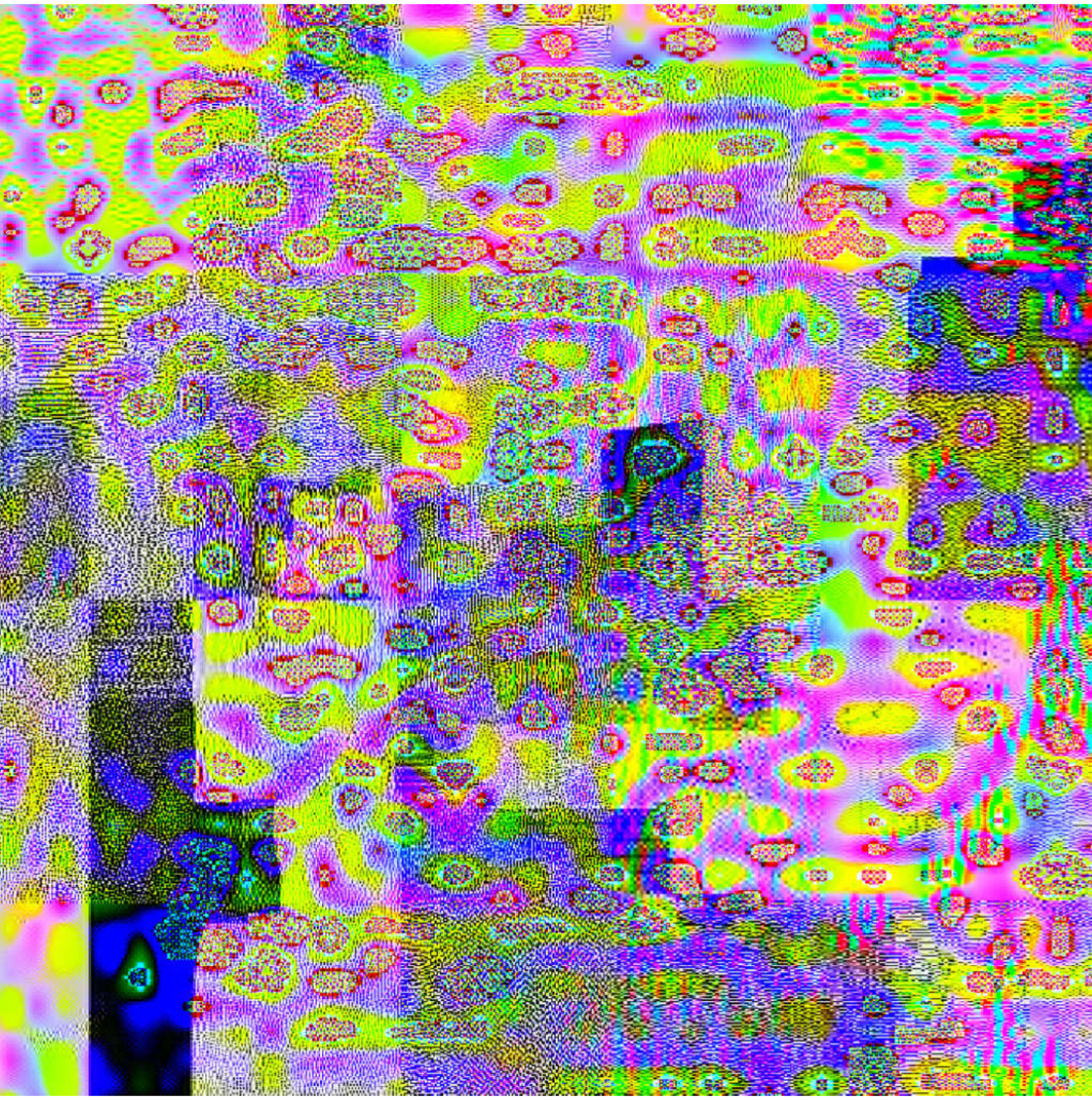}
\caption{Examples of visual corruptions.}
\label{fig:SystemAndCorruptionExamples}
\end{figure}

Visual corruptions are a key symptom of missed GPU defects. Conventional techniques for capturing them rely on humans observing a display and identifying anomalies. This manual process is error-prone, and potentially costly and unscalable.
In our previous work [1], we described a deep learning system that analyzes visual content displayed by the tested GPU and detects anomalies using both supervised and unsupervised learning. Our visual corruption detection system captures the display using a camera (screen grabbing does not enable testing the actual displays used). To control the generation of corruptions (typically quite rare, see Figures \ref{fig:SystemAndCorruptionExamples},\ref{Fig:corruptions}), we reproduced driver bugs that were observed in the past and trained our models on known corruptions. 

While our supervised approach with deep CNN [2] and LSTM [3] topologies obtained high accuracy for known corruptions, model training involved costly manual tagging. Even when reproducing driver bugs, a large portion of video segments may still remain uncorrupted, and manual search for anomalous frames was required. This process led to limited amounts of collected training data, and thus potentially limited exposure and generalization to novel corruptions. Therefore, in addition to supervised algorithms, our system also included unsupervised algorithms to detect anomalies. These algorithms required no manual effort, but suffered from lower accuracy due to lack of supervision.

In this work, we strike a balance between the supervised and unsupervised approaches, using weak supervision models. We make use of the fact that reproducing driver bugs yields visual corruptions with \textit{higher probability} than with running without those malfunctioning drivers. We thus obtain \textit{weak labels}, indicating higher propensity for corruption for videos generated with faulty drivers. Since the process of generating corruptions can be automated, we can all but remove human involvement and potentially obtain unlimited amounts of weakly supervised data.  

We exploit this weak signal with Multiple Instance Learning [4,5] models we adapt to our case. We substantially outperform unsupervised models such as Generalized Adversarial Networks [6], while meeting strict requirements to enable both accurate detection and real-time computational efficiency (each one of our deployed systems processes $2$M frames per day). Finally, in addition to surpassing unsupervised models, we are also able to discover novel visual corruptions that supervised models were unable to detect, due to our ability to scale and generalize.
%While some automatic validation methods exist, they currently typically require exact reference content replayed on the test system, comparing corruptions with expected good outputs. However, due to the great variability and dynamic nature of real-world visual content, expected good outputs cannot be pre-profiled in a scalable, comprehensive way.

%The content in popular computer-games and websites keeps evolving, and also includes randomness, e.g. in the selection of ads or in the probabilistic outcomes of actions in games.

%While there was previous deep-learning work about evaluating natural image quality (see, e.g., [1], [2] and [3]), our setting is different from previous work. It requires classification of both videos and 3D-games, while the content is not known in advance, and millions of images must be processed per day (posing strict requirements on both run-time and false alert rate). Also, the issues to detect are the variety of issues seen in faulty GPUs, not the previously considered camera or communication distortions.

\begin{comment}

\section{System Bckground}
\label{prev-sec}
\input{previouswork.tex}
\end{comment}

\section{Weak Supervision Approach}
\label{weak-sec}
In this section we describe in detail our weak supervision modeling approach, designed to enable both accurate detection and real-time computational efficiency, while removing the need for manual tagging and boosting generalization to novel corruption types.

\textbf{Problem formulation and overview}. Let $\mathcal{X}$ be a set of $N$ training videos $x_1, x_2... x_N$, and let $\mathcal{Y}$ be a set of corresponding \textbf{weak labels} $y_1, x_2... y_N$, with each $y_i \in \{1,0\}$ corresponding to whether video $x_i$ contains a visual corruption or not. $\mathcal{Y}$ are generated automatically by reproducing known bugs as described above, and are dubbed weak labels for two main reasons. First, the labels are \textit{noisy} -- reproducing known bugs does not guarantee visual anomalies, but merely increases the probability of observing them. Second, the labels are \textit{aggregate} -- a video $x_i$ with $y_i = 1$ may potentially be long and nearly devoid of visual corruptions, with only a small segment containing an anomaly. Empirically, we were able to generate corruptions with high enough frequency to ensure our positive videos contain at least one anomaly.   

Importantly, during test-time our real-time system receives short \textbf{segments} of videos (consisting, for example, of $32$ frames). Our goal, based on the above training data with weak labels given for entire videos, is to learn to detect visual corruptions in new incoming \textit{segments} at test time. 

Our approach is based on the Multiple Instance Learning (MIL) setting [4,5,7], where aggregate labels are given for \textit{bags} of instances. A common goal in MIL is to transfer information from labels at the bag level to train an individual-level model [7]. We are interested in going from aggregate video-level labels, to a segment-level model. We model videos as bags of segments, similarly to [4]. 

Our method consists of bag creation, feature extraction, and a weak supervision component. 

\textbf{Bag creation}. 
Let $\mathcal{B}_a$ be a corrupted bag (video), where different continuous segments represent individual instances in the bag, $(p_1, p_2, . . . , p_m)$, where m is the number of segments in the bag. We assume that at least one of these segments contains a corruption. Similarly, uncorrupted videos are represented as normal bags $\mathcal{B}_n$, with continuous segments $(n_1, n_2 , . . . , n_m)$. In the normal bag, none of the instances contain corruptions.

\textbf{Feature extraction}. 
We denote by $f$ a pre-trained C3D [8] trained on the Sports-1M dataset [9] followed by a $3$-layer fully-connected (FC) neural network. During training we freeze the original C3D weights, and update only the FC. 

\textbf{Weak supervision}. We apply a Deep Multiple Instance Learning (MIL) approach as proposed in [4], as well as an MIL Attention method as in [5]. We next elaborate on these two approaches.

\subsection{Deep Multiple Instance Learning}
\begin{figure}[h]
    \centering
    \includegraphics[width=14cm]{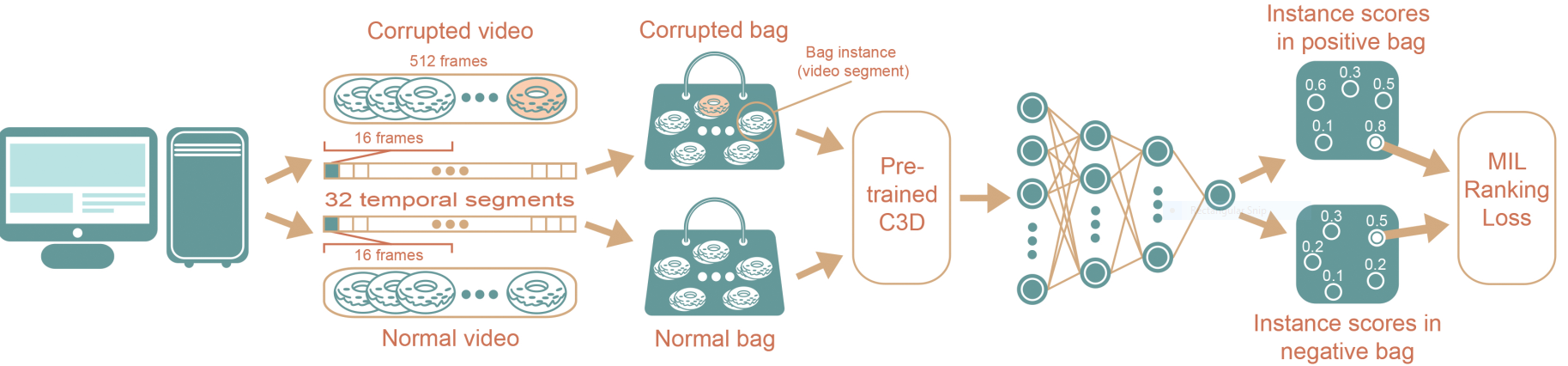}
    \caption{Our Multiple Instance Learning (MIL) model flow requires no manual tagging, using weak labels generated by reproducing known bugs and feeding them into weak supervision models.}
    \label{fig:flow}
\end{figure}

For the loss function we use the following hinge loss $
HL = \max(0,1-\max\limits_{i \in \mathcal{B}_a}f(\mathcal{V}^i_a)+\max\limits_{i \in \mathcal{B}_n}f(\mathcal{V}^i_n))
$, where \(\mathcal{V}^i_a\) is a segment in a corrupted bag ,\(\mathcal{V}^i_n\) is a segment in a normal bag, \(\mathcal{B}_a\) is a corrupted bag and \(\mathcal{B}_n\) is a normal bag. As illustrated in Figure \ref{fig:flow}, for each of $\mathcal{B}_a$ and $\mathcal{B}_n$, we take instances with maximum predicted score under $f$ and penalize pairs of bags where the difference $1-\max\limits_{i \in \mathcal{B}_a}f(\mathcal{V}^i_a)+\max\limits_{i \in \mathcal{B}_n}f(\mathcal{V}^i_n)$ is greater than zero in order to push positive and negative segments apart. We update model weights with mini-batch gradient descent on the objective $\min\limits_{\mathbf{w}}(\frac{1}{z} \sum_{j=1} ^z H\_L) + ||\mathbf{w}||^2$, where $z$ is the batch size and $\mathbf{w}$ are classifier weights.

%\max{(0,1-\mathcal{Y}_\mathcal{B}_j (\max_{i \in \mathcal{B}_j}{(w,\phi(x_i))-b))}}

\subsection{Attention Model}
Rather than taking a hard maximum in our loss, we could instead use a soft weighted average where weights are determined by a neural network as described in [5]. 
Let $H:=\{h_1,...,h_k\}$ be a bag of outputs of the final FC layer of $f$ for each instance (segment) $k$. Then we use the following MIL pooling $z = \sum_{k=1}^K a_k \mathbf{h_k}$, where $a_k = \frac{\exp\{\mathbf{w}^\top \tanh(Vh_k^\top)\}}{\sum_{j=1}^K {\exp\{\mathbf{w}^\top \tanh(Vh_j^\top)\}}}$, $\mathbf{w} \in \mathbb{R}^{L \times 1}$ and $\mathbf{V} \in \mathbb{R}^{L \times M}$, with $M$ the size of the last FC layer, $L$ a hyperparameter.

\section{Experiments and Results}
\label{res-sec}
\textbf{Implementation details}. We use contiguous sequences of $512$ frames as bags, and split them into $32$ contiguous sequences of $16$ frames (segments). We resize each frame to $112\times112\times3$ and finally obtain bags of $32$ segments, each of size $16\times112\times112\times3$.
As in [4], each training batch contains $60$ bags of $30$ corrupted bags and $30$ normal bags. 
Our training data consists of $532$ videos and $216$ for validation. For our test set we have $2573$ uncorrupted videos and $213$ corrupted videos. 

After passing our segments through the pre-trained C3D [8] feature extractor, we obtain a $32\times4096$ output that is passed through a $3$-layer fully-connected neural network. 

Unlike [1,4], we do not assume to be given frame-level labels during validation, hence we make use of a custom accuracy metric on the bag level in which a bag is labeled as positive if at least one of its segments contains a corruption. In addition, we use a tuned threshold to ensure a False Positive Rate (FPR) of $=0.1\%$, a criterion chosen to reflect strict false alert requirements imposed on our system.

More formally, let \( \mathcal{B} \) be a bag, let \( f \) be our trained classifier, let \( i \) be a segment and let \( t\) be a threshold so that the FPR is at most $0.1$\%, i.e. $\max\limits_{t}\{t>0 : \frac{FP(t)}{N}<0.001\}$ (we tune $t$ on $1$M frames of uncorrupted data). Let $\mathcal{A} = \{\mathcal{B} : \mathcal{B} \text{ is a bag which has a corruption in it}\}$ be the set of bags with at least one corrupted segment, and define a bag aggregation function $h_t(\mathcal{B})$ such that $h_t(\mathcal{B}) =1$ if $\max\limits_{i \in \mathcal{B}} f(i) > t $ and $h_t(\mathcal{B}) =0$ otherwise. We then use true positive and true negative sets $\{\mathcal{B}: \mathcal{B} \in \mathcal{A} \land h_t(\mathcal{B})=1\} $ and $\{\mathcal{B}: \mathcal{B} \notin \mathcal{A} \land h_t(\mathcal{B})=0\} $, respectively, to measure accuracy.

For the Deep Multiple Instance model, we trained with dropout of $60\%$, using Adagrad [10] with a learning rate of $0.1$ and epsilon $1e-08$. For the MIL Attention model we trained with $60\%$ dropout using Adam [11]. We pick the best models with a validation set.

\subsection{Comparison to Unsupervised Methods}

Our approach is most comparable to the unsupervised methods reported in [1] that also do not involve any manual tagging: Variational Auto-Encoder (VAE), Generative Adversarial Network (GAN), (un-)normalized energy-based model, and an anomaly detection method [12] on top of C3D features. These models are trained on the $1M$ uncorrupted frames. 

\begin{comment}

\begin{itemize}
	\item \textbf{Variational auto-Encoder (VAE)} A generative model trained by optimizing a variational lower bound to the data likelihood \cite{kingma2013auto} The VAE was trained on videos without corruptions. At inference time, the log-likelihood of the model is used to predict if the input frame is from the same distribution as the training set. The threshold for this decision is chosen based on validation data.
	
	\item \textbf{Generative Adversarial Networks (GANs)} A generative model that learns to model a distribution based on a min-max game \cite{goodfellow2014generative}. 
    We trained the model based on frames without corruptions, and at inference time we used the discriminator's probability to detect corrupted frames. We tried different variants of GANs and selected  the one with the best performance on our validation data - Boundary Equilibrium Generative Adversarial Networks (BEGAN) \cite{began}).

	\item \textbf{Energy-Based model} In this model we split each frame into $32\times32$ pixel patches and calculate the Energy for each patch: $E = \sqrt{\sum{r^2_{i,j,m}}}$ where $r_{i,j,m}$ is the value of the patch at the position $(i,j)$ in the m-th channel, normalized by the patch mean. Then, we calculate the average of the lowest $k$ patches (where $k$ is a hyper-parameter). This average is compared to a threshold obtained on validation data. We normalized each patch to account for recording conditions (dividing by the average energy of the corresponding patches in the three preceding frames).
	
\end{itemize}
\end{comment}

\begin{SCfigure}
    \includegraphics[height=4cm,width=0.4\linewidth]{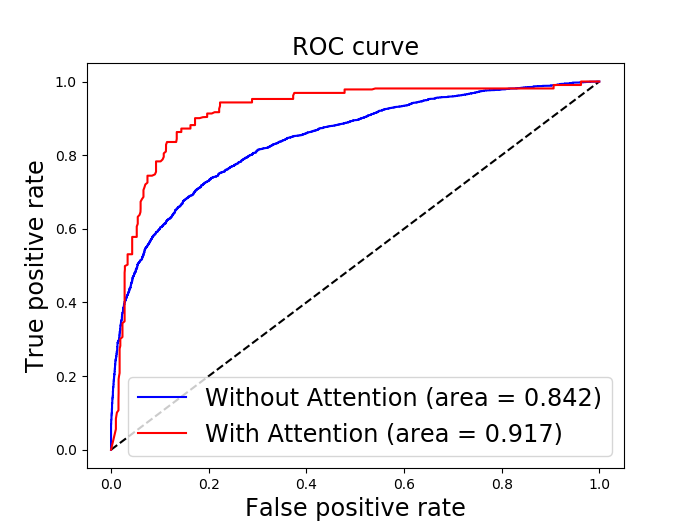}
    \caption{ROC curves for our weak supervision models. The attention-based model is better in terms of overall AUC, but the Deep MIL outperforms it in terms of our Recall@FPR$=0.1$ metric, aimed to reflect our strict False Positive Rate requirements.}
    \label{fig:attentionroc}
\end{SCfigure}

In Table \ref{table:results}, we compare our approach to the unsupervised results of [1] with respect to Recall$@$False Positive Rate$=0.1\%$. We use the Deep MIL approach rather than the attention method as it performed better in terms of this metric. As seen in Figure \ref{fig:attentionroc}, while the attention-based method does better in overall Area Under the Curve (AUC), the Deep MIL method outperforms it in terms of Recall@FPR. 

We significantly surpass the unsupervised approaches by utilizing weak supervision signal, increasing accuracy about $3$-fold from the best baseline (GAN-based). In Table \ref{table:results}, we compare against the GAN model across a range of corruptions and outperform it across $4$ corruption types by a large margin. In our preliminary experiments the unsupervised approach is exposed to a much larger amount of (uncorrputed) data than our model. We expect that by training on larger automatically generated data, our model will be able to surpass the unsupervised approach across more corruptions.

%For the two corruptions where the difference in favor of the GAN approach is more pronounced ("Message popup" and "Macro block"), we observe that the nature of the anomaly does not stray far from the uncorrputed distribution, thus favoring the unsupervised approach which was able to see many more examples from this region in space.}

\begin{table}[ht]
\parbox{.35\linewidth}{
    \centering

\begin{tabular}{lcc}

\toprule
                & Average recall (\%)  \\
\midrule
                Energy w/o norm. & 2.8      \\
                Energy w norm.   & 3.3      \\
                VAE                            & 2.9   \\
                BEGAN                           &   4.7      \\
                C3D-Anomaly                      & 0.7 \\
                Deep MIL                            & \bf{13.0} \\
\bottomrule
            \end{tabular}%
 }
\hfill
\parbox{.55\linewidth}{
  \centering
			\begin{tabular}{lcc}
\toprule
Corruption &Unsupervised & Deep MIL \\
\midrule
Flicker & 4.0 & \bf{13.7} \\
Display stride & 1.2 & \bf{11.4} \\
Vertical, horizontal lines & 4.39 & \bf{27.3} \\
Green flash & 1.5 & \bf{13.1}\\
Color space change & \bf{14.1} & 11.0 \\
Message popup & \bf{11.0} & 5.4 \\
Macro block & \bf{6.3} & 2.7 \\
\bottomrule
			\end{tabular}
				}
\caption {Comparing [1] and our approach. Left: Recall@FPR$=0.1\%$. Right: Results per corruption.}
\label{table:results}
\end{table}

%\begin{wrapfigure}[13]{r}{0.75\textwidth}
\sidecaptionvpos{figure}{c}
%\end{wrapfigure}
\vspace{-10pt}
\subsection{Discovery of New Corruptions}
\begin{figure}[htp]
\centering
\begin{subfigure}{.4\textwidth}
\centering
\includegraphics[width=2cm,height=2cm]{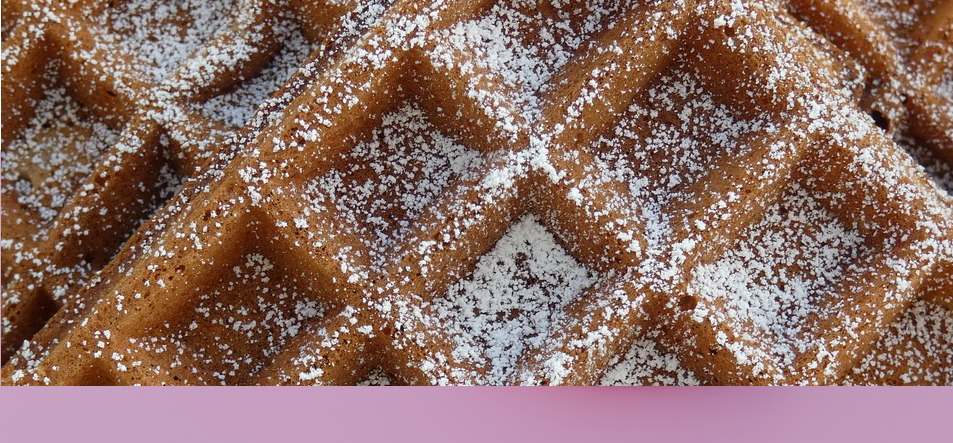}
\includegraphics[width=2cm,height=2cm]{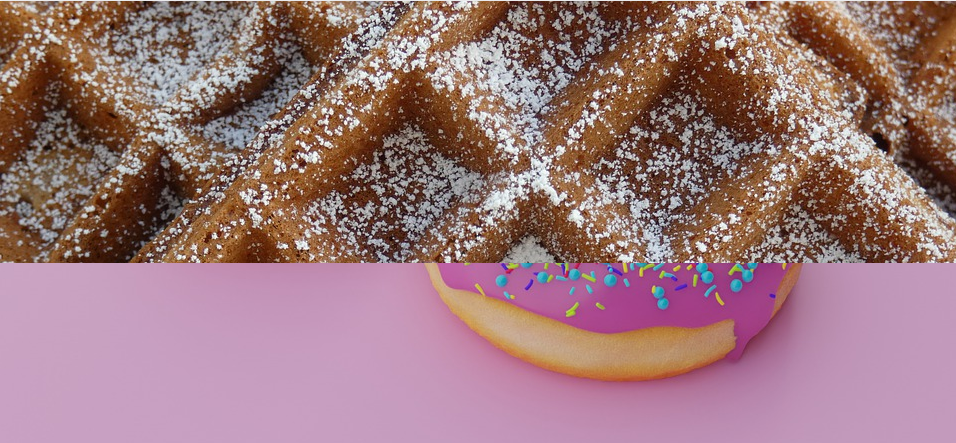}
  \caption{Bottom split, half screen corruptions.}
  \label{Fig:split_examples}
\end{subfigure}%
\begin{subfigure}{.5\textwidth} \hspace{20pt}
\centering
    \includegraphics[width=2cm,height=2cm]{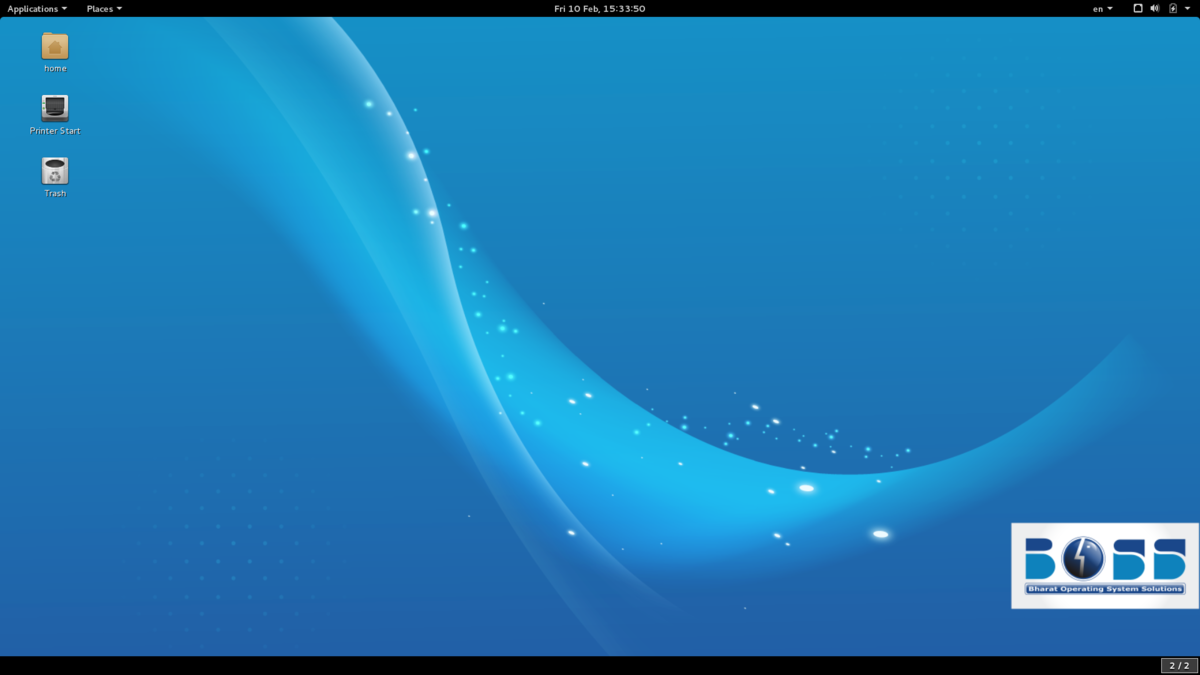}
\includegraphics[width=2cm,height=2cm]{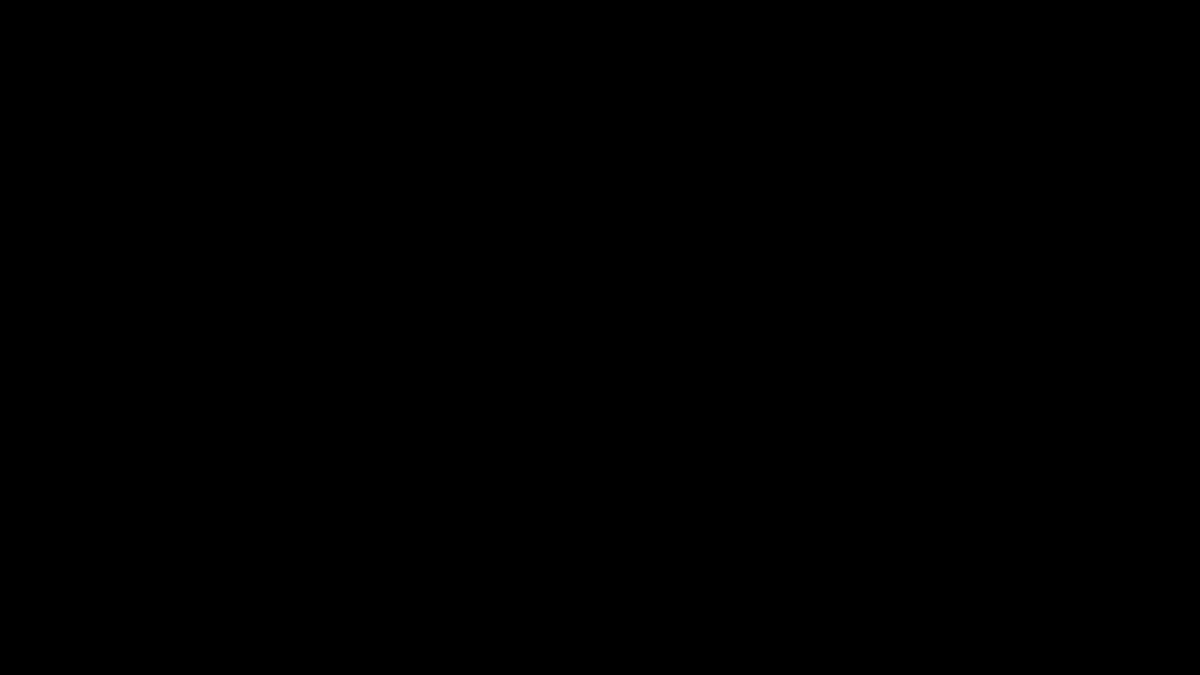}
\includegraphics[width=2cm,height=2cm]{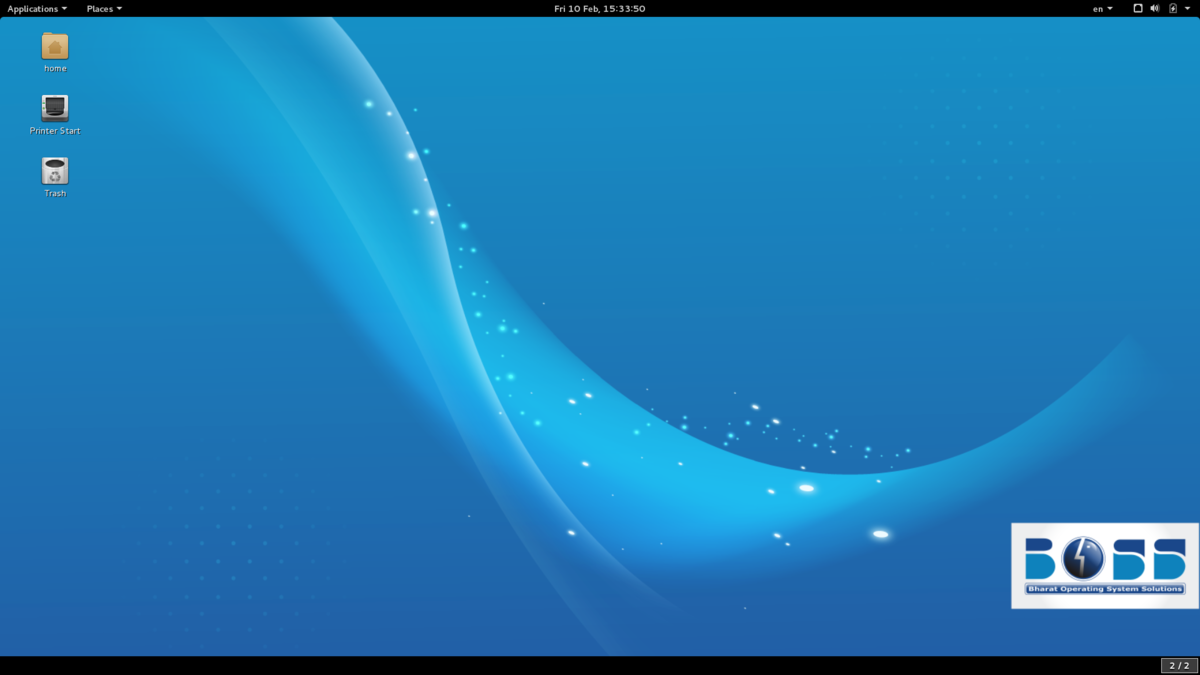}
  \caption{Sequence of sudden blackout corruption. }
  \label{Fig:sudden_black}
\end{subfigure}
\caption{Illustrations of unknown corruptions we discover. Original images from [13,14,15].}
\label{Fig:corruptions}
\end{figure}

We are able to discover new corruptions undetected by both the supervised methods described in [1] as well as the unsupervised baselines. Two of them feature anomalies manifested in a single frame (Figure \ref{Fig:split_examples}). We detected $38$ videos with the \textit{half screen} corruption and $6$ with \textit{bottom split}. 
For both corruptions we achieve a perfect false positive rate. Another type of corruption we discover features a sudden blackout, which is of a more sequential nature (Figure \ref{Fig:sudden_black}).

\section{Discussion and Future Work}
\label{disc-sec}
We described a weak supervision approach for detecting visual corruptions in videos and testing GPUs. Our methods involve no human tagging, outperforms unsupervised approaches and discovers novel corruptions. In the future, we would like to extend our approach to incorporate frame labels when they are given, and also explore the multiclass setting.

%

%\subsubsection*{Acknowledgments}

\section*{References}

\small

[1] Lev Faivishevsky, Ashwin K. Muppalla, Ravid Shwartz-Ziv, Ronen Laperdon, Benjamin Melloul, TahiHollander, and Amitai Armon. Automated testing of graphics units by deep-learning detection of visual anomalies. 2018.

[2] Alex Krizhevsky, Ilya Sutskever, and Geoffrey E Hinton. Imagenet classification with deep convolutional neural networks. In {\it Advances in neural information processing systems}, pages 1097–1105, 2012.

[3] Sepp Hochreiter and Jürgen Schmidhuber. Long short-term memory. {\it Neural computation}, 9(8):1735–1780,1997.

[4] Waqas Sultani, Chen Chen, and Mubarak Shah.  Real-world anomaly detection in surveillance videos. In {\it Proceedings of the IEEE Conference on Computer Vision and Pattern Recognition}, pages 6479–6488,2018.

[5] Maximilian Ilse, Jakub Tomczak, and Max Welling. Attention-based deep multiple instance learning. In {\it International Conference on Machine Learning}, pages 2132–2141, 2018.

[6] David Berthelot, Thomas Schumm, and Luke Metz. Began: boundary equilibrium generative adversarial networks. {\it arXiv preprint arXiv:1703.10717}, 2017.

[7] Dimitrios Kotzias, Misha Denil, Nando De Freitas, and Padhraic Smyth. From group to individual labels using deep features. In {\it Proceedings of the 21th ACM SIGKDD International Conference on Knowledge Discovery and Data Mining}, pages 597–606. ACM, 2015.

[8] Du Tran, Lubomir Bourdev, Rob Fergus, Lorenzo Torresani, and Manohar Paluri. Learning spatiotemporal features with 3d convolutional networks. In {\it Proceedings of the IEEE international conference on computer vision}, pages 4489–4497, 2015.

[9] Andrej Karpathy, George Toderici, Sanketh Shetty, Thomas Leung, Rahul Sukthankar, and Li Fei-Fei. Large-scale video classification with convolutional neural networks. In {\it Proceedings of the IEEE conference on Computer Vision and Pattern Recognition}, pages 1725–1732, 2014.

[10] John Duchi, Elad Hazan, and Yoram Singer.   Adaptive subgradient methods for online learning and stochastic optimization. {\it Journal of Machine Learning Research}, 12(Jul):2121–2159, 2011.

[11] Diederik  P  Kingma  and  Jimmy  Ba.   Adam:  A  method  for  stochastic  optimization. {\it arXiv preprint arXiv:1412.6980}, 2014.

[12] Fei Tony Liu, Kai Ming Ting, and Zhi-Hua Zhou. Isolation-based anomaly detection. {\it ACM Transactionson Knowledge Discovery from Data (TKDD)}, 6(1):3, 2012.

[13] Max Pixel. Sweet Waffle Delicious Food Enjoy Dessert Eat. \url{https://www.maxpixel.net/Sweet-Waffle-Delicious-Food-Enjoy-Dessert-Eat-4477521}.

[14] Max Pixel. Donut Frosting Food Sugar Bakery Eat Bake.  \url{https://www.maxpixel.net/Donut-Frosting-Food-Sugar-Bakery-Eat-Bake-4475458}.

[15] Wikimedia Commons. A basic desktop environment of BOSS Linux 6.1 on a personal computer with menus above and a taskbar below. \url{https://commons.wikimedia.org/wiki/File:Screenshot\_of\_BOSS\_Linux\_v6.1\_Desktop\_Environment.png}, 2017.

%\bibliographystyle{unsrt}
%\bibliography{ref}

\end{document}